\PassOptionsToPackage{table}{xcolor}
\documentclass[runningheads]{llncs}

\usepackage{main}
\usepackage{mainabbrv}
\usepackage{graphicx}
\usepackage{booktabs}
\usepackage[accsupp]{axessibility}
\usepackage[pagebackref,breaklinks,colorlinks]{hyperref}
\usepackage{orcidlink}

\usepackage{multirow}
\usepackage{algpseudocode}
\usepackage{algorithm}
\usepackage{algorithmicx}
\usepackage{csquotes}

\begin{document}

\title{SeSame: Simple, Easy 3D Object Detection with Point-Wise Semantics} 

\titlerunning{SeSame: Simple, Easy 3D Object Detection with Point-Wise Semantics}

\author{Hayeon O\inst{1}\and
        Chanuk Yang\inst{2}\and
        Kunsoo Huh\inst{2}\orcidlink{0000-0002-7179-7841}}

\authorrunning{H.O, C.Yang, K.Huh}

\institute{%
  \begin{tabular}{c}
    Department of Automotive Engineering (Automotive-Computer Convergence) \\
    Department of Automotive Engineering \\
    Hanyang University, Seoul, 04763, Republic of Korea \\ 
    \texttt{\{gkdus9595, ych901208, khuh2\}@hanyang.ac.kr}
  \end{tabular}%
}

\maketitle

\begin{abstract}
In autonomous driving, 3D object detection provides more precise information for downstream tasks, including path planning and motion estimation, compared to 2D object detection. In this paper, we propose SeSame: a method aimed at enhancing semantic information in existing LiDAR-only based 3D object detection. This addresses the limitation of existing 3D detectors, which primarily focus on object presence and classification, thus lacking in capturing relationships between elemental units that constitute the data, akin to semantic segmentation. Experiments demonstrate the effectiveness of our method with performance improvements on the KITTI object detection benchmark. Our code is available at \href{https://github.com/HAMA-DL-dev/SeSame}{https://github.com/HAMA-DL-dev/SeSame}

\end{abstract}

\section{Introduction}
\label{sec:intro}
The autonomous driving system is structured in a sequential flow where preceding tasks influence subsequent ones. As the perception system is the initial step in identifying the surrounding environment for autonomous driving, it bears the responsibility of providing accurate information to downstream tasks, e.g. decision-making, path planning, and control. Therefore, various tasks for recognizing driving situations using cameras and LiDAR have been proposed. Notably, 2D detection based on image from camera has achieved accuracy surpassing human identification capabilities due to the rapid advancement of deep learning and ongoing research. However, it has limitations in conveying height and shape information about objects, hindering the delivery of precise information. For this reason, the importance of 3D object detection has emerged. Moreover, according to \cite{survey:vision-centricBEV, survey:surround-view}, this provides advantages in representing 3D object detection results in the Bird's-Eye View (BEV), where the BEV plane acts as the global coordinate for downstream tasks in autonomous driving. In other words, accurate results from 3D object detection also enhance the precision of downstream tasks. For these reasons, 3D object detection is indispensable in autonomous driving. 
For 3D object detection in autonomous driving, image-based methods estimate depth and generate pseudo-LiDAR point clouds\cite{pseduo-lidar}. Nevertheless, this approach tends to have lower accuracy compared to LiDAR-only 3D object detection, and employing multi-view images to enhance precision increases significant computation cost. Consequently, other fusion methods\cite{PointPainting, F-PointNet, EPNet, MV3D, BEVFusion, AVOD,CLOCs} have been proposed. But, the multi-modality method presents challenges in integrating and synchronizing data, which have different representations and are defined in different coordinate systems. On the other hand, LiDAR-only based 3D object detection has the advantage of fully leveraging the geometric information provided by point cloud. Furthermore, LiDAR semantic segmentation\cite{SqueezeSeg,SalsaNet} has demonstrated the ability to extract sufficient semantic information from point clouds. This led to a question: 
\begin{quotation}
   "what if we can extract sufficient semantic features from point cloud and leverage them in existing 3D detectors?"
\end{quotation}

Therefore, inspired by the question and \cite{PointPainting}, this paper proposes a method to utilize semantic information obtained from LiDAR semantic segmentation for LiDAR-only 3D object detection, aiming to minimize loss and distortion on geometric information and preserve it. This method consists of three components, where LiDAR semantic segmentation provides supplementary semantic features for 3D detector. In summary, the contributions of this paper are as follows: 

\begin{enumerate}
\item We propose a simple and easy method for 3D object detectors by integrating segmented point cloud obtained from LiDAR semantic segmentation. To our best knowledge, it is the first to leverage semantic segmentation to object detection.
\item  Unlike existing methods that rely on calibration, this approach does not require any calibration, thus minimizing loss and distortion in extracting and incorporating semantic information.
\item The proposed method outperforms multi-modality approach for all classes, and the reference model and the baseline detector for car on KITTI object detection benchmark.
\end{enumerate}

\section{Related Works}
\subsection{LiDAR Semantic Segmentation}
\begin{figure}[ht!] 
  \centering
  \centerline{\includegraphics[width=12.0cm]{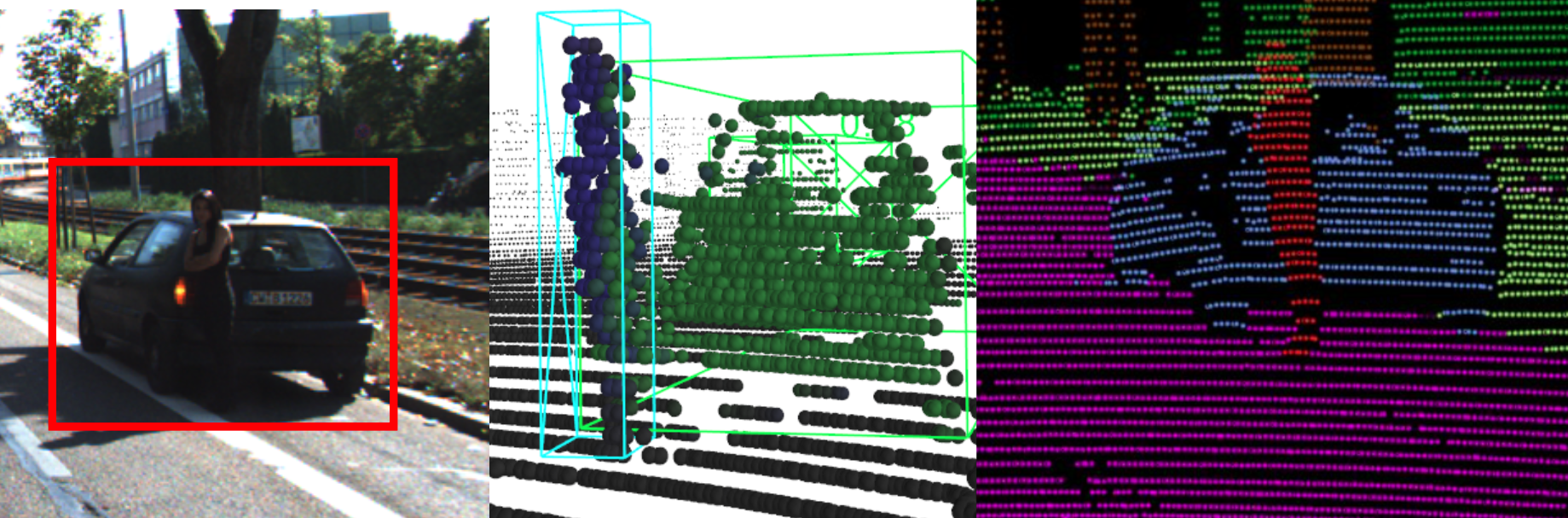}}
  \caption{\textbf{(left)} This depicts a scenario in which two objects, pedestrian and car, are overlapping, causing occlusion. \textbf{(center)} For the pixel-wise segmentation\cite{deeplabv3} projected onto the point cloud using perspective projection,,miss-segmentation occurs in which some of the semantic features of the pedestrian (blue) are classified as car (green). \textbf{(right)} On the other hand, it may be seen that point-wise semantic segmentation has higher accuracy.}
  \label{fig:sem_seg_difference}
\end{figure}

To fully understand the driving scenario in autonomous driving, precise semantic segmentation is required. Depending on camera and LiDAR, semantic segmentation can be divided into 2D semantic segmentation at the pixel-wise level and 3D semantic segmentation at the point-wise level\cite{survey:semantic-segmentation}. The tasks perform multi-class classification on pixels and points that constitute image and point cloud, respectively. Originated from the PointNet series\cite{PointNet,PointNet++}; they are for indoor space, limiting their ability to extract semantic features from outdoor driving environments which have diverse density contexts. Thus, other methods are proposed to guarantee semantic features from point cloud. Inspired by \cite{FCN, UNet, SqueezeNet},  SqueezeSeg\cite{SqueezeSeg} and PointSeg\cite{PointSeg} utilize spherical projection to convert point cloud into range-view image, apply 2D semantic segmentation, and then map the result back to the point cloud. However, spherical projection has the drawback of causing loss or distortion of semantic information from the camera and geometric information from the point cloud. Z.Zhuang \etal \cite{PMF} address this limitation with two-stream network and perspective projection. Nevertheless, similar to \cref{fig:sem_seg_difference}, this method is sensitive to the calibration, thus its slight variations can significantly affect the results\cite{TransFusion}. On the other hand, inspired by \cite{UNet}, Y.Zhang \etal\cite{PolarNet} propose BEV projection method to project point cloud onto a 2D grid. Unlike voxelization defined in Cartesian coordinate\cite{VoxelNet}, this uses Polar coordinate system to encode point clouds. However, this method also cannot fully preserve geometric information because it projects 3D point clouds into 2D space. Cylinder3D\cite{Cylinder3D}, instead of other projection methods, converts point clouds defined in Cartesian coordinate into Cylindrical coordinate, considering the sparsity of point cloud. Its asymmetrical residual block considers the characteristic of many objects in autonomous driving scenes being cuboid-shaped. And the dimension-decomposition considers various contexts that point clouds represent in outdoor spaces with different densities. To reflect these high-rank contexts, they are divided according to context, enabling a LiDAR semantic segmentation that fully understands high-rank contexts with low computation cost. For this reason, we select Cylinder3D\cite{Cylinder3D} to extract the semantic information for 3D detectors. 

\subsection{3D Object Detection with Images} 
\label{sec:3ddet_with_image}
This can be divided into mono, stereo, and multi-view image methods depending on whether they use single or multiple images. Wang \etal\cite{pseduo-lidar} generate depth map from estimated depth. This is utilized as a point cloud, providing geometric information lacking in raw image, and is used to LiDAR-based detection. D.Park \etal\cite{DD3D} proposes a fully convolutional single-stage 3D object detection that combines the pseudo-LiDAR and end-to-end method with mono image as input and \cite{FPN} as the backbone with three tasks, predicting logit for each class, 2D boxes, and 3D boxes. BEVFormer\cite{BEVFormer} extracts BEV features from multi-view images. Its spatial cross-attention between multi-camera views connects the same objects across different camera views and extract spatial information as a grid-shaped BEV query. However, due to the fundamental characteristics of image, inaccurate estimated depth can interfere with detection performance. Furthermore, this method can suffer from efficiency issues compared to LiDAR-based detection as concatenating multiple images increases the input dimension, leading to higher computational cost and lower performance. \\

\begin{figure}[ht!] 
  \centering
  \centerline{\includegraphics[width=12.0cm]{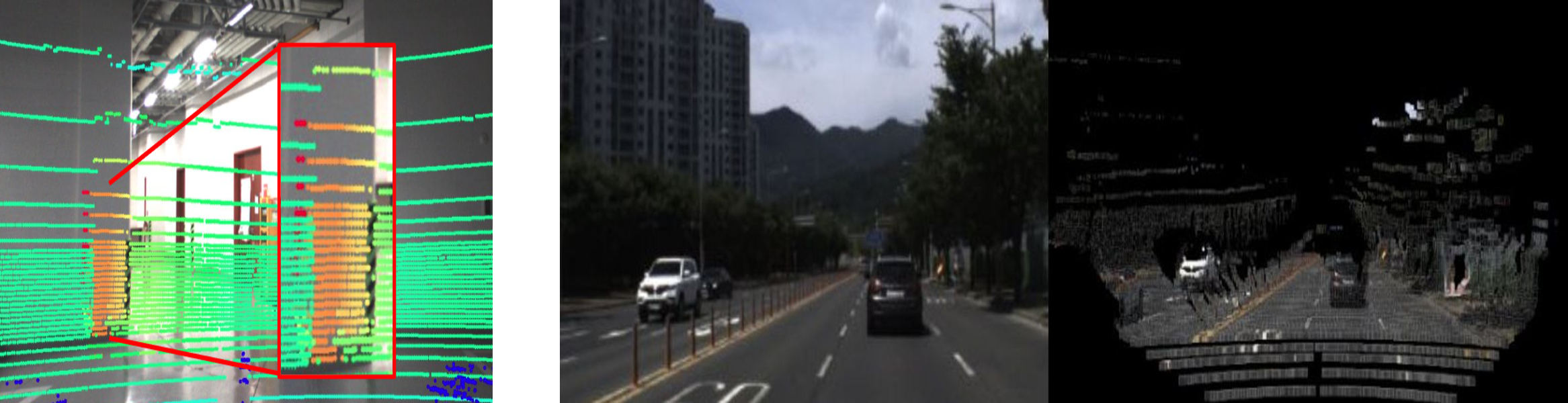}}
  \caption{\textbf{(left)}Due to the error-sensitive hard association based on the LiDAR-camera calibration matrix, some points were not projected onto reflective surfaces. \textbf{(right)}Additionally, many image features did not correspond to the point cloud. it is stated that less than 5\% of image features match with point clouds (for a 32-channel LiDAR scanner)\cite{BEVFusion}}.
  \label{fig:lidar2cam_and_cam2lidar}
\end{figure}

Based on above, multi-modal 3D object detection has been proposed. In \cite{PointPainting}, pixel-wise semantic feature from 2D semantic segmentation is concatenated with raw point cloud and applied to 3D object detection, aiming to supplement semantic feature lacking in raw point cloud. Z.Liu \etal\cite{BEVFusion} proposed a method where both features are represented in a unified representation on BEV and performs both 3D object detection and BEV map segmentation based on fused features. AVOD\cite{AVOD} and MV3D\cite{MV3D} aggregate features extracted from images and point cloud to generate 3D proposals. CLOCs\cite{CLOCs} aims to improve detection performance by fusing 2D bounding boxes obtained from 2D detector and 3D bounding boxes obtained from 3D detector. However, these multi-modal methods pose additional challenges, as it requires considering various representations defined in different domain as a unified representation and demands synchronization for actual implementation. And it requires the calibration matrix between LiDAR and camera to consider unified representation. If there is even slight error in this matrix, this can significantly impact performance as mentioned in \cite{TransFusion}. And also in \cref{fig:lidar2cam_and_cam2lidar}, semantic loss can occur due to an insufficient number of point clouds matching pixels. \\

\subsection{3D Object Detection with Point Cloud}
\label{sec:3ddet_with_point_cloud}
\noindent
PointRCNN\cite{PointRCNN} is a two-stage detector composed of region proposal, classification, and box regression. It uses PointNet++\cite{PointNet++} as the backbone network to obtain point-wise feature vectors from an encoder and decoder. These vectors are utilized for 3D proposal generation and foreground point segmentation, resulting in semantic features and foreground masks, respectively. Merged them with raw point cloud, region pooling returns semantic features and local spatial points. Finally, 3D boxes of detected objects are returned after refinement and confidence prediction. SECOND\cite{SECOND}, which utilizes voxel grouping for point cloud on a 3D grid, is inspired by VoxelNet\cite{VoxelNet}. Unlike the 3D dense convolutional neural network in \cite{VoxelNet}, the sparse convolutional neural network in \cite{SECOND} considers only non-zero data and optimizes convolution operations through fast rule generation for repetitive patterns. This optimization greatly enhances speed during training and inference. On the other hand, PointPillars\cite{PointPillar} considers point cloud defined in 2D grid along with the corresponding feature map, enabling it to apply only 2D convolutions for highly efficient operations. However, these methods primarily focus on detecting the presence of objects in the scene, leading to limitation in obtaining context information between points. Therefore, LiDAR-only based 3D object detection has inconsistency between localization and classification of 3D bounding boxes\cite{EPNet}.

\section{Method}
\begin{figure}[htb!] 
  \centering
  \centerline{\includegraphics[width=12.0cm]{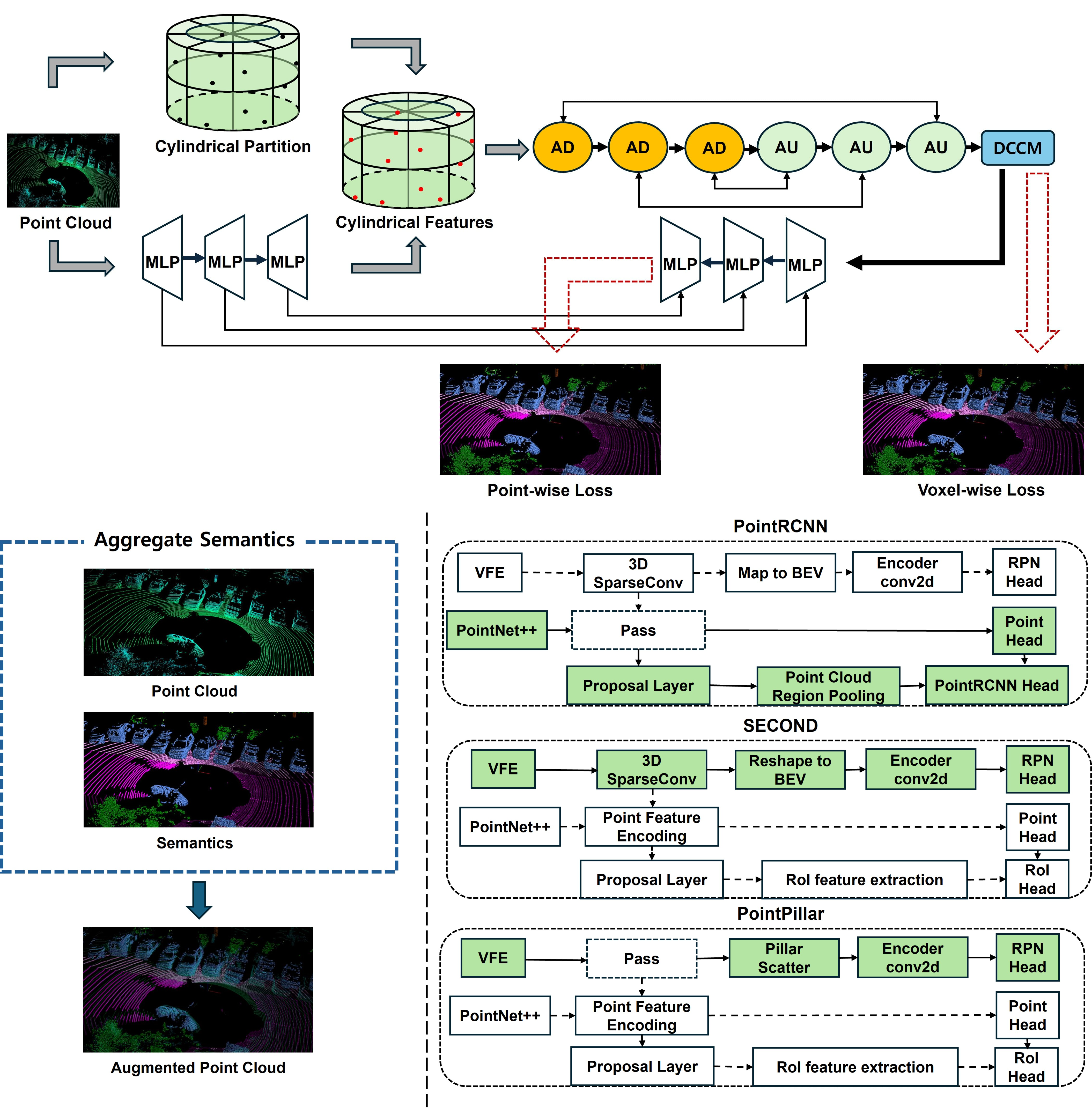}}
  \caption{\textbf{(up)} overall architecture of LiDAR sem.seg.\cite{Cylinder3D} implemented in this paper. It returns per point semantics. \textbf{(down-left)} The input point clouds are augmented with the semantics. \textbf{(down-right)} 3D detectors with various feature extractors ; point, voxel, pillar. OpenPCDet\cite{framework:pcdet} supports various models within a single framework, with the green sections indicating the components used by each model.}
  \label{fig:overview}
\end{figure}

\subsection{Extract Semantics from Point Cloud}
\noindent
The LiDAR semantic segmentation in this paper requires a point cloud as input and outputs per-point class labels. This can be summarized as a feature extraction method for point clouds similar to the approach in \cite{PointPainting}. However, unlike 2D semantic segmentation, LiDAR semantic segmentation has its advantages. In 2D semantic segmentation, semantic features obtained from images are fused with the point cloud using a camera-LiDAR calibration matrix through perspective projection. Any slight error in this matrix can result in inaccurate projection of the 2D semantic features onto the point cloud. Furthermore, in autonomous driving and robotics, cameras and LiDAR often operate at different frequencies, causing latency issues. In contrast, the LiDAR semantic segmentation used in this paper considers only the point cloud, eliminating the need for calibration and synchronization issues between data with different frames. For these reasons, we select LiDAR semantic segmentation to extracts semantic information from the point cloud. Among the models available, we chose the method proposed by Zhu Xinge \etal\cite{Cylinder3D}, which introduces Cylindrical coordinates to better preserve the 3D information of the point cloud while accounting for its sparsity.

\subsection{Aggregate the Semantics}
\noindent
\textbf{Motivation.}
We were highly motivated by \cite{PointPainting}. In the referenced paper, semantic features obtained from stereo images were matched and then concatenated with the point cloud. These features were mapped to three classes at the pixel-wise level, derived from a 2D semantic segmentation \cite{deeplabv3+}. However, this matching was based on perspective projection, which can cause loss and distortion as mentioned earlier. Therefore, we utilize semantic information directly extracted from the point cloud, preserving 3D geometric information without such projection. \\

\noindent
\textbf{Proposed Method.} 
As shown in \cref{alg:semantic_feature_concatenation}, the semantics from previous step include an information per each point, indicating which class it belongs to. This corresponds to the semantic label map of the SemanticKITTI\cite{dataset:semanticKITTI}, which is the training dataset for the pretrained model from the previous step. Therefore, class mapping is required for the detector training on the KITTI 3D object detection dataset. Among the semantics, only classes corresponding to the KITTI label map are assigned, followed by one-hot encoding. After aggregating with the raw point cloud, a new point cloud is generated as follows : [$x_i, y_i, z_i, r_i, {\text{semantics}}$], where $i$ is index number, $x, y, z$ are coordinates of a point in Cartesian coordinates, and $r$ is intensity. Here, ${\text{semantics}}$ represent semantic information for unlabeled, car, pedestrian, and cyclist, respectively. The point cloud with semantics obtained as follows becomes the input of feature extractor of each detectors.

\begin{algorithm}[ht!]
\caption{Pseudocode of Aggregate Semantics}
\label{alg:semantic_feature_concatenation}
\begin{algorithmic}
\renewcommand{\algorithmicrequire}{\textbf{Input:}}
\renewcommand{\algorithmicensure}{\textbf{Output:}}

\Require{
    \\ map: returns index corresponding to the class \\
    point cloud: $P = \{p_i=(x_i, y_i, z_i, r_i) \mid i = 1, \ldots, n\}$  $\in \mathbb{R}^{N \times 4}$ with $N$ points \\ 
    semantics: generated from LiDAR sem. seg.  $\in \mathbb{R}^{N \times C}$ with $C$ classes
}
\Ensure{
    \\ augmented point cloud : $P_L = \{p_i\textsuperscript{L}=(x_i, y_i, z_i, r_i, semantics) \,|\, i = 1, \ldots, n\}$ $\in \mathbb{R}^{N \times (C+4)}$\\
}

\State \texttt{def augment\_points (point\_cloud, map, semantics) $:$ }

\State \texttt{    \hspace{\algorithmicindent}label\_one\_hot = np \. zeros((points.shape[0], len(map))) }
\State \texttt{    \hspace{\algorithmicindent}for index, point in point\_clouds$:$}
\State \texttt{        \hspace{\algorithmicindent}\hspace{\algorithmicindent} if semantics[i] in map$:$}
\State \texttt{        \hspace{\algorithmicindent} \hspace{\algorithmicindent}\hspace{\algorithmicindent}label\_one\_hot[index][map[cls]] += 1.0}
\State \texttt{    \hspace{\algorithmicindent}label\_one\_hot = torch.from\_numpy(label\_one\_hot).float()}
\State \texttt{    \hspace{\algorithmicindent}augmented\_point\_cloud = np.concatenate((points,label\_one\_hot),axis=1)}
\State \texttt{    \hspace{\algorithmicindent}return augmented\_point\_cloud}
\end{algorithmic}
\end{algorithm}

\subsection{LiDAR Detectors}
To evaluate the proposed method, we selected networks among existing 3D detectors that take point cloud as input. As mentioned in \cite{PointPainting}, these networks vary based on how they encode the point cloud. \cite{PointRCNN} uses raw input points without separate voxelization to extract point-wise features. In contrast, \cite{SECOND,PointPillar} voxelizes the input point cloud into voxels and pillars, respectively. Therefore, their 3D feature extraction backbone (VFE) generates 3D features, which are then mapped to Bird's Eye View (BEV) and serve as input to the 2D backbone. \cite{PointRCNN}, excluding this process. Each network's head predicts bounding boxes using features extracted from their backbone.

\subsection{Loss Functions}

\textbf{LiDAR Semantic Segmentation.}
For network optimization, weighted cross-entropy loss is used for voxel-wise and point-wise loss to maximizing per point accuracy and IoU score for all classes and Lovasz-softmax~\cite{loss:lovasz-softmax} loss is used for point-wise loss to supervise the training. They share the same weight. So, the total loss is a summation of them. 

\begin{equation}
  L_{total} = L_{voxel}+L_{point}
  \label{eq:important}
\end{equation}

\noindent
\textbf{LiDAR Detector.}
\begin{itemize}
    \item \textbf{SeSame+point.} 
    The loss function is defined in two stages:
    
    In stage 1, focal loss is used to address class-imbalance because background points, obtained through foreground segmentation, are more numerous than foreground points. For bin-based classification in 3D box generation, cross-entropy loss is used for the \( x \) and \( z \) axes and orientation, while Smooth L1 loss is used for regression based on residual values for the \( y \) axis and object dimensions \( (h, w, l) \). 
    \begin{equation}
    L_{\text{reg}} = \text{smooth}_{L1} \sum_{v_i \in \{\Delta y, h, w, l\}} (v_i, v_i^*)
    \end{equation}
    
    where \(\Delta y\) represents the residual values for the \( y \) axis, and \(h, w, l\) are the object dimensions. In stage 2, a positive label is assigned if the Intersection over Union (IoU) is 0.55 or higher; otherwise, it is considered a negative label. Only proposals with positive labels are refined, and cross-entropy loss is computed. So the total loss function is then a combination of these losses:
    
    \begin{equation}
    L_{\text{total}} = L_{\text{focal}} + L_{\text{cls}} + L_{\text{reg}} + L_{\text{refine}}
    \end{equation}

    \item \textbf{SeSame+voxel and SeSame+pillar.} Ground truth and anchor boxes are defined as $(x, y, z, w, h, l, \theta)$. Localization loss is calculated through Smooth L1 loss for the residual between these two boxes. The probability associated with each anchor serves as an argument for the focal loss used to calculating the classification loss. Additionally, directional loss is added to these two losses to compute the final loss.
    \begin{equation}
      L_{\text{total}} = L_{\text{loc}} + L_{\text{cls}}
    \end{equation}
\end{itemize}

\section{Experiment}
\textbf{Training Details.}
SeSame+point, +voxel, and +pillar are trained with learning rates of 0.01, 0.003, and 0.003, and batch sizes of 8, 16, and 16, respectively. All three detectors share the same Adam as optimizer and OneCycleLR as scheduler\cite{adam_onecyle}, with a weight decay rate of 0.01 and momentum of 0.9, and are trained for 80 epochs with a single TITAN RTX GPU. \\

\noindent
\textbf{SemanticKITTI.} Cylinder3D was trained on the SemanticKITTI dataset, which consists of point clouds classified into 28 classes. They are merged into a total of 19 classes by combining classes based on different moving statuses and ignoring classes with too few points. The label data containing this information is of \text{\texttt{uint32\_t}}, where the upper half represents the instance label and the lower half represents the semantic label. In this paper, only the lower half containing semantic labels from this data was used. The 19 classes  were mapped to three classes defined in the KITTI object detection benchmark dataset: car, pedestrian, and cyclist. \\

\noindent
\textbf{KITTI 3D Object Detection Evaluation.} 
The dataset contains 7481 training samples and 7518 testing samples, consisting of images and LiDAR point clouds. All existing detectors\cite{PointRCNN,SECOND,PointPillar} use the dataset, but differ in data split. In this paper, following general split method, the given training dataset was further split into 3712 training and 3769 validation samples for experiment. The train and val samples obtained through this split do not overlap\cite{NIPS2015_6da37dd3}. Before the splitting, pretrained Cylinder3D is used to perform inference on this dataset, and the resulting label mapped to the KITTI dataset is concatenated with training sample, thus forming the training dataset. The same approach was applied to the test split, and the results were submitted to the official test server, showing the outcomes as in \cref{table:results_on_kitti_3ddet,table:results_on_kitti_bevdet}.The evaluation metric used was Average Precision (AP), with IoU thresholds of 0.7 for cars and 0.5 for pedestrians and cyclists, respectively. The three classes are divided into easy, moderate, and hard based on detection difficulty, where occlusion and truncation increase from the former to the latter. \\

\noindent
\textbf{Data Augmentation.}
Due to the limited number of training samples in the dataset, common data augmentation techniques applied in existing detectors are incorporated. Following the approach initiated from \cite{VoxelNet}, a database is first created from the given training dataset and its labels. Random selection is made from this database for training, involving data augmentation such as random flipping along the x-axis, rotation around the z-axis and the origin within the range of $[-\pi/4, \pi/4]$, and random scaling within the range of $[0.95, 1.05]$. These augmentations are applied to both the parameters of individual bounding boxes and the point sets within those boxes, being configured differently for each detector. To fairly evaluate the effectiveness of the proposed method, common settings are referenced and applied uniformly. If the 3D bounding boxes and point clouds generated by this technique result in physically implausible scenarios, such as collision due to overlapping bounding boxes, the original data is reverted.

\subsection{Overall Results}

\textbf{Quantitative Results}
The results are based on the evaluation detection metrics which are: BEV, 3D as shown in \cref{table:results_on_kitti_3ddet,table:results_on_kitti_bevdet}. The modalities are LiDAR(L), and camera(C). The proposed method is applied to three detectors classified based on the type of input features, outperforming both the camera-only method and the multi-modality method. Among these detectors, when using point feature as input, it outperforms the baseline detector on car. This result is evident in both metrics, which signifies the consistency of the results. As mentioned in \cite{PointPillar}, increasing per box data augmentation leads to a further degradation in performance for pedestrians. This was also observed in our result. A comparison with before and after the application of the proposed method, can be seen in \cref{sec:ablation_study}.

\begin{table}[htb!]
\caption{Results on the KITTI test 3D object detection benchmark}
\label{table:results_on_kitti_3ddet}
\resizebox{\textwidth}{!}{%
\begin{tabular}{|c|c|ccc|ccc|ccc|}
\hline
\multirow{2}{*}{Method} & \multirow{2}{*}{modality} & \multicolumn{3}{c|}{Car} & \multicolumn{3}{c|}{Pedestrian} & \multicolumn{3}{c|}{Cyclist} \\
 &  & Easy & Mod. & Hard & Easy & Mod. & Hard & Easy & Mod. & Hard \\ \hline
DD3D\cite{DD3D} & C (mono) & 23.22 & 16.34 & 14.20 & 13.91 & 9.30 & 8.05 & 2.39 & 1.52 & 1.31 \\
Pseudo-LiDAR\cite{pseduo-lidar} + AVOD\cite{AVOD} & C (stereo) & 55.4 & 37.2 & 31.4 & N/A & N/A & N/A & N/A & N/A & N/A \\
MV3D\cite{MV3D} & L + C & 71.09 & 62.35 & 55.12 & N/A & N/A & N/A & N/A & N/A & N/A \\
AVOD-FPN\cite{AVOD} & L + C & 73.59 & 65.78 & 58.38 & 38.28 & 31.51 & 26.98 & 60.11 & 44.90 & 38.80 \\
PI-RCNN\cite{PI-RCNN} & L + C & 84.37 & 74.82 & 70.03 & N/A & N/A & N/A & N/A & N/A & N/A \\
F-PointNet\cite{F-PointNet} & L + C & 82.19 & 69.79 & 60.59 & 50.53 & 42.15 & 38.08 & 72.27 & 56.12 & 49.01 \\
PointRCNN\cite{PointRCNN} & L (point) & \textbf{85.94} & 75.76 & 68.32 & 49.43 & 41.78 & 38.63 & 73.93 & \textbf{59.60} & \textbf{53.59} \\
SECOND\cite{SECOND} & L (voxel) & 83.13 & 73.66 & 66.20 & 51.07 & 42.56 & 37.29 & 70.51 & 53.85 & 46.90 \\
PointPillars\cite{PointPillar} & L (pillar) & 79.05 & 74.99 & 68.30 & \textbf{52.08} & \textbf{43.53} & \textbf{41.49} & \textbf{75.78} & 59.07 & 52.92 \\ \hline
SeSame + point \textbf{(Ours)} & L (point) & 85.25 & \textbf{76.83} & \textbf{71.60} & 42.29 & 35.34 & 33.02 & 69.55 & 54.56 & 48.34 \\
SeSame + voxel \textbf{(Ours)} & L (voxel) & 81.51 & 75.05 & 70.53 & 46.53 & 37.37 & 33.56 & 70.97 & 54.36 & 48.66 \\
SeSame + pillar \textbf{(Ours)} & L (pillar) & 83.88 & 73.85 & 68.65 & 37.61 & 31.00 & 28.86 & 64.55 & 51.74 & 46.13 \\ \hline
\end{tabular}%
}
\end{table}

\begin{table}[htb!]
\caption{Results on the KITTI test BEV detection benchmark}
\label{table:results_on_kitti_bevdet}
\resizebox{\textwidth}{!}{%
\begin{tabular}{|c|c|ccc|ccc|ccc|}
\hline
\multirow{2}{*}{Method} & \multirow{2}{*}{modality} & \multicolumn{3}{c|}{Car} & \multicolumn{3}{c|}{Pedestrian} & \multicolumn{3}{c|}{Cyclist} \\
 &  & Easy & Mod. & Hard & Easy & Mod. & Hard & Easy & Mod. & Hard \\ \hline
DD3D \cite{DD3D} & C (mono) & 30.98 & 22.56 & 20.03 & 15.90 & 10.85 & 8.05 & 3.20 & 1.99 & 1.79 \\
Pseudo-LiDAR \cite{pseduo-lidar} + AVOD \cite{AVOD} & C(stereo) & 66.8 & 47.2 & 40.3 & N/A & N/A & N/A & N/A & N/A & N/A \\
MV3D \cite{MV3D} & L + C & 86.02 & 76.90 & 68.49 & N/A & N/A & N/A & N/A & N/A & N/A \\
AVOD-FPN \cite{AVOD} & L + C & 88.53 & 83.79 & 77.90 & \textbf{58.75} & \textbf{51.05} & \textbf{47.54} & 68.06 & 57.48 & 50.77 \\
PI-RCNN \cite{PI-RCNN} & L + C & 91.44 & 85.81 & 81.00 & N/A & N/A & N/A & N/A & N/A & N/A \\
F-PointNet \cite{F-PointNet} & L + C & 91.17 & 84.67 & 74.77 & 57.13 & 49.57 & 45.48 & 77.26 & 61.37 & 53.78 \\
PointRCNN \cite{PointRCNN} & L(point) & \textbf{92.13} & 87.39 & 82.72 & 54.77 & 46.13 & 42.84 & \textbf{82.56} & \textbf{67.24} & \textbf{60.28} \\
SECOND \cite{SECOND} & L(voxel) & 89.39 & 83.77 & 78.59 & 55.99 & 45.02 & 40.93 & 76.50 & 56.05 & 49.45 \\
PointPillars \cite{PointPillar} & L(pillar) & 90.07 & 86.56 & 82.81 & 57.60 & 48.64 & 45.78 & 79.90 & 62.73 & 55.58 \\ \hline
\textbf{SeSame + point (Ours)} & L(point) & 90.84 & \textbf{87.49} & \textbf{83.77} & 48.25 & 41.22 & 39.18 & 75.73 & 61.70 & 55.27 \\
\textbf{SeSame + voxel (Ours)} & L(voxel) & 89.86 & 85.62 & 80.95 & 50.12 & 41.59 & 37.79 & 76.95 & 59.36 & 53.14 \\
\textbf{SeSame + pillar (Ours)} & L(pillar) & 90.61 & 86.88 & 81.93 & 44.21 & 37.31 & 35.17 & 72.22 & 60.21 & 53.67 \\ \hline
\end{tabular}%
}
\end{table}

\begin{figure}[htb!]
  \centering
  \centerline{\includegraphics[width=12.0cm]{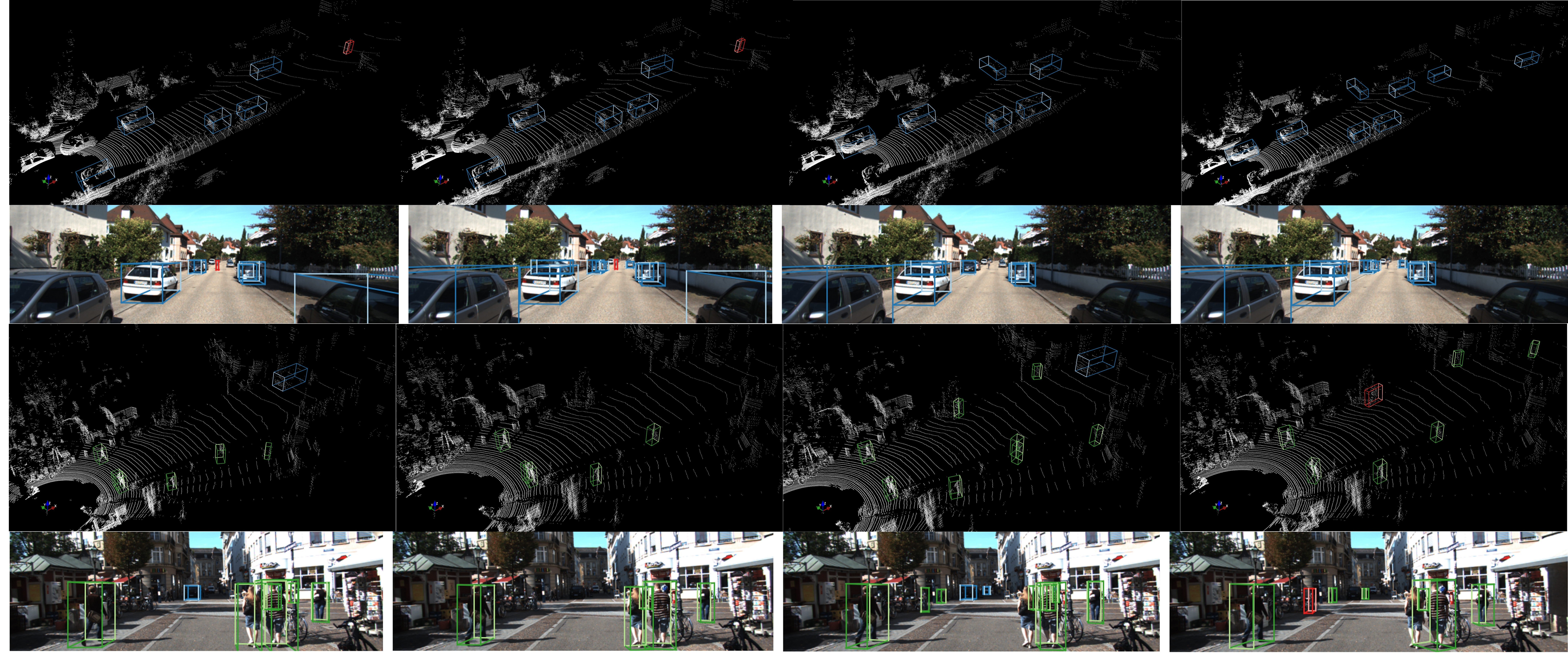}}
  \caption{The leftmost of the four sections represents the ground truth(GT), while the remaining three sections depict predictions from detectors based on input features of point\cite{PointRCNN}, voxel\cite{SECOND}, and pillar\cite{PointPillar}. The top figure illustrates a scenario with multiple cars\textbf{(blue)} and some cyclists\textbf{(red)}, while the bottom figure shows multiple pedestrians\textbf{(green)}.}
  \label{fig:3ddet_qualitative}
\end{figure}

\begin{figure}[htb!]
  \centering
  \centerline{\includegraphics[width=12.0cm]{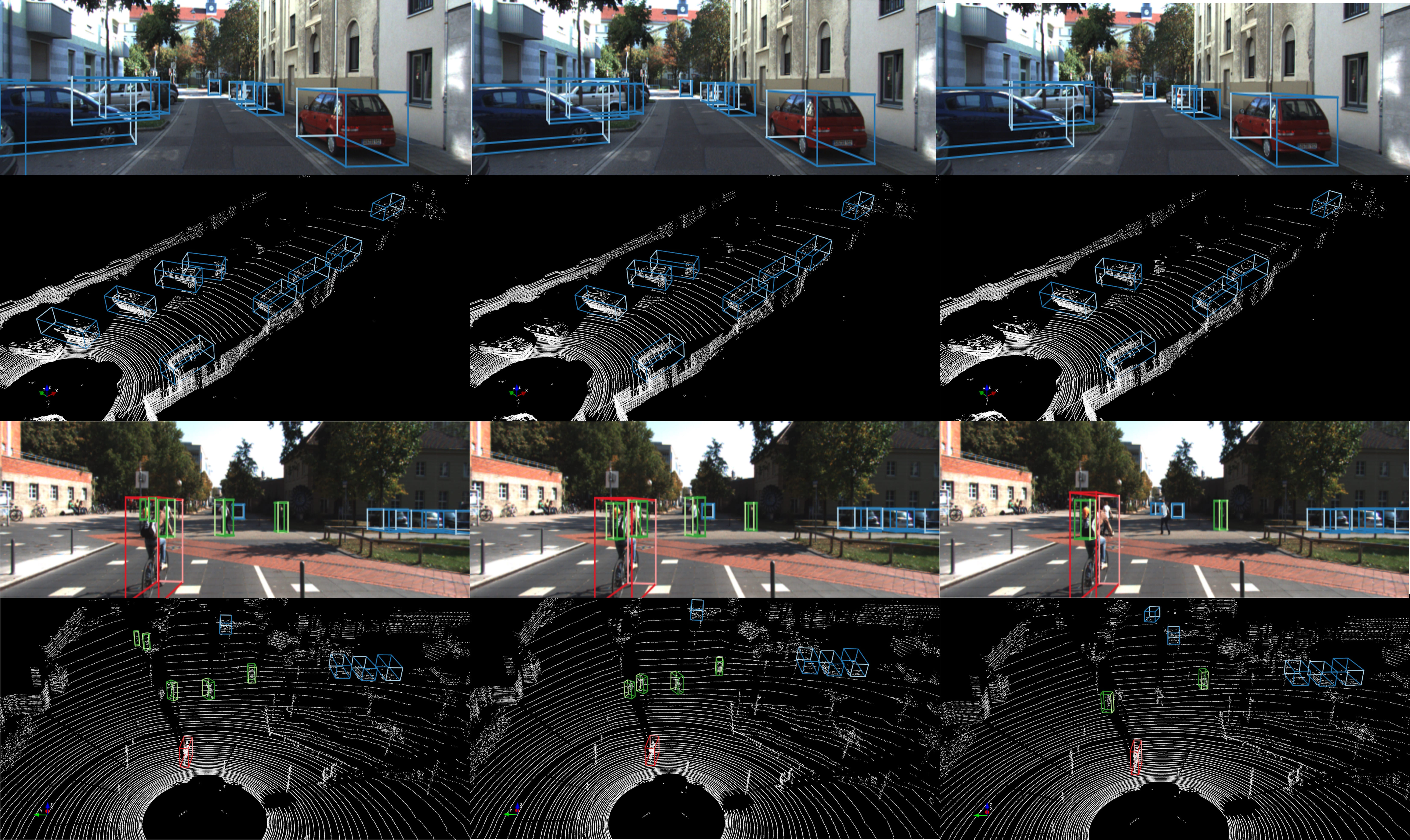}}
  \caption{Qualitative results on the KITTI test set. There are two scenes. For each scene, the results of SeSame+point, +voxel, and +pillar are shown from leftmost to rightmost.}
  \label{fig:3ddet_qualitative_testsplit}
\end{figure}

\noindent
\textbf{Qualitative Results}
As shown in \cref{fig:3ddet_qualitative} and \cref{fig:3ddet_qualitative_testsplit}, SeSame+point detects cars more than the other approaches, which corresponded to the actual cars present in the scene. While SeSame+voxel and And SeSame+voxel recognizes pedestrian better than the other two approaches, which aligns with the quantitative results presented in \cref{table:results_on_kitti_3ddet} and \cref{table:results_on_kitti_bevdet}. On the other hand,  SeSame+point and SeSame+pillar seem to have false positive and false negative for pedestrian each other. However, they also seem to detect other objects correctly, and especially +pillar detects far away car, which was confirmed to be actual objects upon inspection of the scene. Overall, SeSame+point exhibits excellent performance, although false negative occurs in sparse point regions. SeSame+voxel and +pillar effectively identifies objects even in sparse point cloud, with +voxel identifying cars missed by SeSame+point. These findings are consistent with the quantitative results in the paper, demonstrating the overall result is reliable.

\vspace{-\baselineskip} 

\subsection{Ablation Study}
\label{sec:ablation_study}

\textbf{Encoding : Label and Score.}
The label obtained from LiDAR semantic segmentation is mapped from SemanticKITTI to KITTI, assigning a score vector composed of 0 or 1 according to one-hot encoding to the point cloud. When comparing the feature concatenated point cloud with the point cloud passed through softmax for score mapping, the accuracy improved for the score mapping with val split. However, for test split, the accuracy was higher when mapped by one-hot encoding label. Furthermore, there was a discrepancy in performance obtained from the val split and the test split when mapped by score. This implies that the training was within the range of noise for score mapping, and mapping by label is more reasonable. The corresponding results are included in Table 1 in the Supplementary Material.

\noindent
\textbf{Modality : point cloud and multimodal}
As mentioned earlier in \cref{sec:3ddet_with_image}, the multimodal methods that utilize both image and point cloud necessitate calibration, as illustrated in \cref{fig:lidar2cam_and_cam2lidar}, which can distort or lose semantic information from images or geometric information from point clouds. Our method aims to address these challenges and demonstrates superior performance without calibration compared to multimodal methods as shown in \cref{table:comparison_reference_and_ours_test}. This indicates that the proposed method can be free from calibration-induced distortions and losses, while showcasing superior performance even without images.

\noindent
\textbf{Before and After Application of Proposed Method.}
\label{sec:ablation_study_before_and_after_proposed_method}
When comparing detectors using our method with their base models, the results are as follows: they all show performance improvements for cars, especially with SECOND using voxel features as input showing the most benefit as in \cref{table:comparision_existing_and_ours}. For SeSame+voxel, performance gains are observed not only for car but also for cyclist. But \cite{PointPillar} did not show gains in identifying pedestrians and cyclist. We initially assumed that this was caused by PointPillar's conversion of 3D features into 2D features. However, this hypothesis was rejected because SECOND\cite{SECOND} did not cause any performance degradation despite also including the conversion to 2D features. This can be attributed to the stochastic sampling approach employed by PointPillars'\cite{PointPillar} on the point cloud. This sampling substantially filter out or eliminate points containing semantic information, especially when the ground truth object consists of only a few points. Thus, there can be degradation of performance. 

\begin{table}[ht!]
\caption{We compared the performance before and after applying the proposed method based on the KITTI object detection benchmark for car with IoU threshold 0.7. Detectors with the method showed performance improvement.}
\label{table:comparision_existing_and_ours}
\centering
\resizebox{\textwidth}{!}{%
\begin{tabular}{|c|c|c|ccc|ccc|}
\hline
 &  &  & \multicolumn{3}{c|}{\textbf{AP\(_{3D}\)}} & \multicolumn{3}{c|}{\textbf{AP\(_{BEV}\)}} \\
\multirow{-2}{*}{Method} & \multirow{-2}{*}{Modality} & \multirow{-2}{*}{\begin{tabular}[c]{@{}c@{}}mAP\\ (Hard)\end{tabular}} & Easy & Mod. & Hard & Easy & Mod. & Hard. \\ \hline
PointRCNN \cite{PointRCNN} & L(point) & 76.71 & 86.96 & 75.64 & 70.70 & 92.13 & 87.39 & 82.72 \\
SeSame+Point \textbf{(Ours)} & L (point) & \textbf{77.69} & 85.25 & 76.83 & 71.60 & 90.84 & 87.49 & 83.77 \\
\rowcolor[HTML]{FFFFC7} 
Improvement &  &  & -1.71 & \textbf{1.19} & \textbf{0.9} & -1.29 & \textbf{0.1} & \textbf{1.05} \\ \hline
SECOND \cite{SECOND} & L (voxel) & 72.08 & 83.13 & 73.66 & 66.20 & 88.07 & 79.37 & 77.95 \\
SeSame+Voxel \textbf{(Ours)} & L (voxel) & \textbf{75.74} & 81.51 & 75.05 & 70.53 & 89.86 & 85.62 & 80.95 \\
\rowcolor[HTML]{FFFFC7} 
Improvement &  &  & -1.62 & \textbf{1.39} & \textbf{4.33} & \textbf{1.79} & \textbf{6.25} & \textbf{3.00} \\ \hline
PointPillars \cite{PointPillar} & L (pillar) & 74.07 & 79.05 & 74.99 & 68.30 & 88.35 & 86.10 & 79.83 \\
SeSame+Pillar \textbf{(Ours)} & L (pillar) & \textbf{75.29} & 83.88 & 73.85 & 68.65 & 90.61 & 86.88 & 81.93 \\
\rowcolor[HTML]{FFFFC7} 
Improvement &  &  & \textbf{4.83} & -1.14 & \textbf{0.35} & \textbf{2.26} & \textbf{0.78} & \textbf{2.10} \\ \hline
\end{tabular}%
}
\end{table}

\noindent
\textbf{Comparison between Reference Method and Ours.} 
\label{sec:ablation_study_reference_and_ours}
\noindent
There is only one result of \cite{PointPainting} on the test split, painted PointRCNN, so we compared it with SeSame+point as shown in \cref{table:comparison_reference_and_ours_test}. PointPainting\cite{PointPainting} showed decrease in performance and our method exhibited superior performance improvement for car. However, it performed better for pedestrians and cyclists. We hypothesize that the observed results are due to the differences in per-class performance between pixel-wise and point-wise semantic segmentation. \cite{PointPainting} extracts semantic information using DeepLabv3+\cite{deeplabv3+}, trained on Cityscapes\cite{dataset:cityscapes}, achieving IoU of 87.95 and 78.88 for pedestrian and cyclist classes, respectively. In contrast, the point-wise semantic segmentation\cite{Cylinder3D} used in this paper achieves IoU of 73.90 and 65.80 for pedestrian and cyclist classes, respectively. We can see \cref{table:comparison_baseline_reference_and_ours_on_val} supports this hypothesis. This indicates that 3D object detection using semantic information is dependent on the performance of semantic segmentation. Therefore, while LiDAR-based semantic segmentation can provide reliable semantic information for classes where it performs well, it may lead to performance degradation for classes where it does not. This leads us to focus on future work in \cref{sec:conclusion}.

\begin{table*}[htb!]
\caption{Performance comparison of BEV and 3D object detection with reference method and ours for the car class on the test split. The IoU threshold is 0.7. The mAP (Mod.) column represents the mean Average Precision at the moderate difficulty level, providing a balanced assessment of model performance.}
\label{table:comparison_reference_and_ours_test}
\centering
\resizebox{\textwidth}{!}{%
\begin{tabular}{|c|c|c|ccc|ccc|}
\hline
\multirow{2}{*}{Method} & \multirow{2}{*}{Modality} & \multirow{2}{*}{\begin{tabular}[c]{@{}c@{}}mAP\\ (Mod.)\end{tabular}} & \multicolumn{3}{c|}{AP\(_{3D}\)} & \multicolumn{3}{c|}{AP\(_{BEV}\)} \\
 &  &  & Easy & Mod. & Hard & Easy & Mod. & Hard \\ \hline
MV3D & L+C & 69.63 & 71.09 & 62.35 & 55.12 & 86.02 & 76.90 & 68.49 \\
AVOD-FPN & L+C & 74.86 & 73.59 & 65.78 & 58.38 & 88.53 & 83.79 & 77.90 \\
PI-RCNN & L+C & 80.32 & 84.37 & 74.82 & 70.03 & 91.44 & 85.81 & 81.00 \\
F-PointNet & L+C & 77.23 & 82.19 & 69.79 & 60.59 & 91.17 & 84.67 & 74.77 \\ \hline
Painted PointRCNN\cite{PointPainting} & L+C & 79.91 & 82.11 & 71.70 & 67.08 & \textbf{92.45} & \textbf{88.11} & 83.36 \\
SeSame+Point \textbf{(Ours)} & L (point) & \textbf{82.16} & \textbf{85.25} & \textbf{76.83} & \textbf{71.60} & 90.84 & 87.49 & \textbf{83.77} \\ \hline
\end{tabular}%
}
\end{table*}

\begin{table}[htb!]
\caption{Performance comparison of 3D object detection with baseline, reference method, and ours on KITTI val split.}
\label{table:comparison_baseline_reference_and_ours_on_val}
\resizebox{\textwidth}{!}{%
\begin{tabular}{|c|c|lll|lll|lll|}
\hline
\multirow{2}{*}{Method} & \multirow{2}{*}{Modality} & \multicolumn{3}{c|}{Car} & \multicolumn{3}{c|}{Pedestrian} & \multicolumn{3}{c|}{Cyclist} \\
 &  & \multicolumn{1}{c}{Easy} & \multicolumn{1}{c}{Mod.} & \multicolumn{1}{c|}{Hard} & \multicolumn{1}{c}{Easy} & \multicolumn{1}{c}{Mod.} & \multicolumn{1}{c|}{Hard} & \multicolumn{1}{c}{Easy} & \multicolumn{1}{c}{Mod.} & \multicolumn{1}{c|}{Hard} \\ \hline
PointRCNN & L(point) & \multicolumn{1}{c}{86.75} & \multicolumn{1}{c}{76.05} & \multicolumn{1}{c|}{74.30} & \multicolumn{1}{c}{63.29} & \multicolumn{1}{c}{58.32} & \multicolumn{1}{c|}{51.59} & 83.68 & 66.67 & 61.92 \\
Painted PointRCNN & L+C & \multicolumn{1}{c}{88.38} & \multicolumn{1}{c}{77.74} & \multicolumn{1}{c|}{76.76} & \multicolumn{1}{c}{\textbf{69.38}} & \multicolumn{1}{c}{\textbf{61.67}} & \multicolumn{1}{c|}{\textbf{54.58}} & 85.21 & 71.62 & 66.98 \\
SeSame + point \textbf{(ours)} & L(point) & \textbf{88.76} & \textbf{78.35} & \textbf{77.43} & 62.83 & 55.30 & 51.35 & \textbf{87.80} & \textbf{72.74} & \textbf{67.00} \\ \hline
SECOND & L(voxel) & 86.85 & 76.64 & 74.41 & 67.79 & 59.84 & 52.38 & 84.92 & 64.89 & 58.59 \\
Painted SECOND & L+C & 87.15 & 76.66 & 74.75 & \textbf{68.57} & \textbf{60.93} & \textbf{54.01} & \textbf{85.61} & \textbf{66.44} & \textbf{64.15} \\
SeSame + voxel \textbf{(ours)} & L(voxel) & \textbf{88.35} & \textbf{78.55} & \textbf{77.27} & 56.02 & 52.37 & 48.34 & 81.34 & 65.97 & 61.16 \\ \hline
PointPillars & L(pillar) & \textbf{87.22} & \textbf{76.95} & 73.52 & 65.37 & 60.66 & 56.51 & \textbf{82.29} & 63.26 & 59.82 \\
Painted PointPillars & L+C & 86.26 & 76.77 & 70.25 & \textbf{71.50} & \textbf{66.15} & \textbf{61.03} & 79.12 & \textbf{64.18} & \textbf{60.79} \\
SeSame + pillar \textbf{(ours)} & L(pillar) & 85.65 & 75.94 & \textbf{73.85} & 53.43 & 48.43 & 44.69 & 77.85 & 61.45 & 58.32 \\ \hline
\end{tabular}%
}
\end{table}

\subsection{More analysis on : image vs. point cloud for semantics}
\label{sec:limitation}
In this section, we analyze why the proposed method is effective for certain classes but not for others. Improvements were observed for the car class, but not for pedestrians and cyclists as mentioned in \cref{sec:ablation_study_reference_and_ours}. Conversely, when compared to the method base on 2D semantic segmentation \cite{PointPainting}, its performance declined for car while it improved for pedestrian and cyclist as shown in \cref{table:comparison_baseline_reference_and_ours_on_val}. Analyzing the results from both modalities, it is evident that images outperform point clouds in extracting semantic information for objects expected to have a sparse point cloud, such as pedestrians and cyclists. On the other hand, in the case of larger objects like car, occlusion or truncation may occur in images, but the point cloud can identify regions even where occlusion and truncation occur in the image. Therefore, LiDAR semantic segmentation and 2D semantic segmentation each has the advantage of accurately capturing semantics even in environment with occlusion and truncation, effectively capturing the semantics of objects that are small in size and difficult to identify through point cloud alone. For example, we can see this that SeSame+point successfully captured vehicles obscured by overlapping structures as illustrated in \cref{fig:3ddet_qualitative}, and this supports the earlier assertion that point clouds are more robust to occlusion and truncation than images.

\section{Conclusion}
\label{sec:conclusion}
In this paper, we explored the utilization of semantic information obtained from LiDAR semantic segmentation for LiDAR-only 3D object detection, examining its potential impact. This was validated by observing performance improvements of the baseline model on the KITTI 3D object detection benchmark. Furthermore, we demonstrated superior performance compared to multimodal methods that require calibration, which can induce loss of semantic features in images. However, both our method and the reference model suffer from the limitation of requiring data annotation for semantic segmentation, thus incurring the costs associated with labeling. Based on these findings, our future research focuses on self-supervised multimodal semantic segmentation as pretext task for 3D object detection, which leverages both modalities without requiring any data annotation.

%
%
\bibliographystyle{splncs04}
\bibliography{main}
\end{document}